\documentclass[10pt]{article}
\usepackage[a4paper,textwidth=18cm,textheight=24cm,top=2.85cm, bottom=2.85cm, left=1.5cm, right=1.5cm]{geometry}

\usepackage{icdp2009}

\makeatletter
\long\def\@makecaption#1#2{%
\vskip\abovecaptionskip
\sbox\@tempboxa{#1. #2}%
\ifdim \wd\@tempboxa >\hsize
#1. #2\par
\else
\global \@minipagefalse
\hb@xt@\hsize{\box\@tempboxa\hfil}%
\fi
\vskip\belowcaptionskip}
\makeatother

\usepackage{lmodern}

\usepackage{graphicx}
\usepackage{times}
\usepackage{amssymb}
\usepackage{amsbsy}
\usepackage{amsmath}
\usepackage{amsthm} % This is only for theorem and proofs.
\usepackage{setspace}
\usepackage{ifpdf}
\usepackage{titlesec}
\usepackage{array}
\usepackage[bottom]{footmisc}
\usepackage{booktabs}
\usepackage{hhline}
\usepackage{enumitem}
\usepackage{float}
\usepackage{comment}
\usepackage[font=small, labelfont=bf]{caption}
\usepackage{subcaption}
\usepackage{multirow}
\usepackage[linktoc=all]{hyperref}
\usepackage{xcolor}
\begin{document}
\noindent

\newcommand\sara[2]{\textcolour{#1}{#2}}
\bibliographystyle{IEEEtran}

\title{Domain Adversarial Training for Infrared-colour Person Re-Identification}

\authorname{Nima Mohammadi Meshky, Sara Iodice, Krystian Mikolajczyk}
\authoraddr{Imperial College London, United Kingdom}

\newcommand{\norm}[1]{\left\lVert #1 \right\rVert}
\newcommand{\ith}{$i^\text{th}$}
\newcommand{\cth}{$c^\text{th}$}
\newcommand{\lth}{$l^\text{th}$}
\newcommand{\jth}{$j^\text{th}$}

\titlespacing*{\section}{0pt}{1.5ex plus 0ex minus 0ex}{1.5ex plus 0ex}
\titlespacing*{\subsection}{0pt}{1.25ex plus 0ex minus 0ex}{1.25ex plus 0ex}
\titlespacing*{\subsubsection}{0pt}{1ex plus 0ex minus 0ex}{1ex plus 0ex}

\maketitle


\abstract
Person re-identification (re-ID) is a very active area of research in computer vision, due to the role it plays in video surveillance. Currently, most methods only address the task of matching between colour images. However, in poorly-lit environments CCTV cameras switch to infrared imaging, hence developing a system which can correctly perform matching between infrared and colour images is a necessity. In this paper, we propose a part-feature extraction network to better focus on subtle, unique signatures on the person which are visible across both infrared and colour modalities. To train the model we propose a novel variant of the domain adversarial feature-learning framework~\cite{ganin2016domain}. Through extensive experimentation, we show that our approach outperforms state-of-the-art methods.

\section{Introduction}
\begin{figure}[!t]
\centering
\includegraphics[width=0.99\linewidth, height=1.7in]{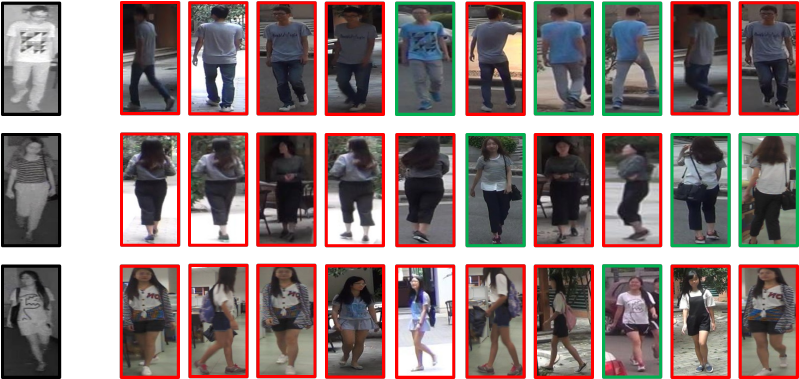}
\caption{Examples of rank-list given by ResNet50  from the SYSU-MM01 dataset. We rank all gallery pictures by the distance to the probe images (from left to right). Frames with the same IDs are green-framed.} 
\label{fig1}
\vspace{-\baselineskip}
\end{figure}

When being presented with an image containing a person of interest, person re-ID tells us whether the person has been observed at another place from a different camera. More specifically, person re-ID can be formulated as the task of ranking a set of stored images (gallery set) based on similarity to a probe image.
The task is non-trivial due to the discrepancies which arise from capturing a given person across different camera views. One practical application is tracking a missing person in densely populated urban scenarios such as a theme park, university campus, shopping mall, etc. The significance in both research and application makes person re-ID a popular topic in computer vision. Currently, the majority of works in the literature consider the problem of matching among colour images only. However, in current-generation surveillance systems, CCTV cameras switch to infrared imaging in poorly lit environments, hence developing methods which can additionally deal with the discrepancies associated with infrared-colour matching is crucial. What makes infrared-colour person re-ID more challenging than the conventional setting is that colour information cannot be utilized. In figure~\ref{fig1} we show some examples of rank-lists given by our ResNet50 baseline from the SYSU-MM01 dataset~\cite{wu2017rgb} (images with the same IDs are green-framed). It can be noticed that if any one of the probe images had been captured in colour, it would be easier to correctly match it to the relevant images in the gallery set due to the unique colour of clothing which the person of interest is wearing. However, in absence of such information ambiguity is introduced, as non-relevant identities in the gallery set who have body-structure and clothing akin to the person of interest can now appear equally or even more similar to the probe image. To address this issue, we utilize a novel variant of the domain adversarial learning framework~\cite{ganin2016domain} to extract a feature representation that is invariant across colour and infrared modalities. Furthermore, in contrast to existing solutions~\cite{dai2018cross,wang2019learning} considering only the global representation from the ResNet model, we leverage part-level features with a proposed variant of the PCB model~\cite{sun2018beyond}. We believe the proposed model is better equipped to focus on subtle features on a persons clothing such as a t-shirt logo, shirt collar, folded-up sleeve, etc., which otherwise would be missed. Such local signatures on a person are both unique and present across both colour and infrared modalities.

\begin{figure*}[!t]
\includegraphics[width=1\textwidth]{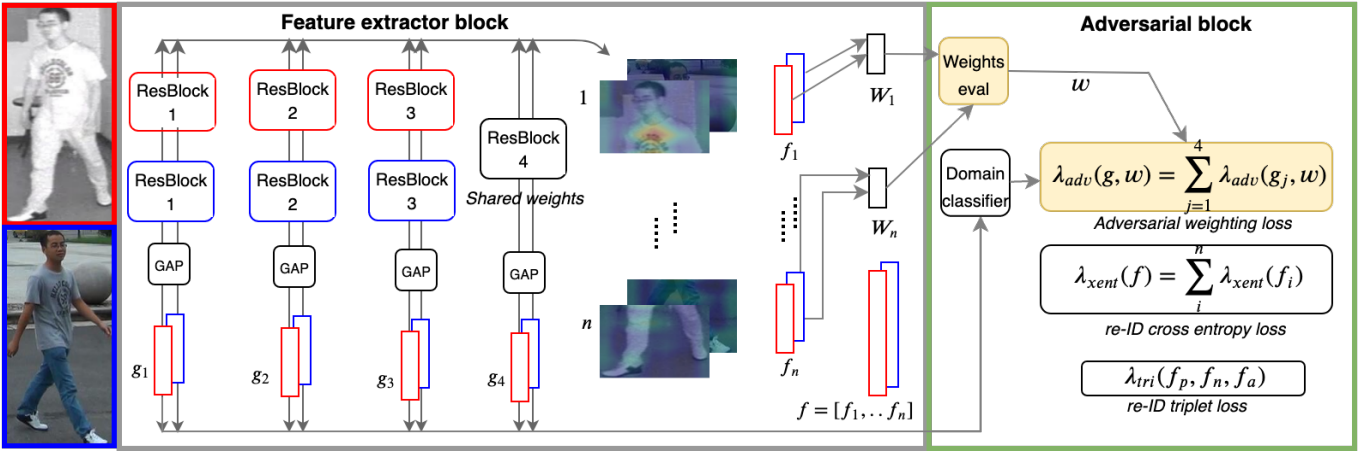}
\caption {The proposed framework with two components. The feature extractor block  extracts features at multiple abstraction levels from intermediate layers of ResNet50 backbone, i.e., $g_j: j=1..4$ and  part-level features from the last feature map, i.e., $f_i: i=1..n$. The adversarial block reduces the domain discrepancy between RGB and infrared modalities in the feature space combining a domain classifier with our weighting strategy.}
\label{diagram_model}
\vspace{-\baselineskip}
\end{figure*}

\vspace{0.5\baselineskip}
\noindent Our main  contributions are summarised below.

\begin{itemize}
    \item We propose a novel variant of the domain adversarial loss function~\cite{ganin2016domain}, which a) operates on intermediate layers to address the domain discrepancy on low-level features where matching characteristics across the modalities are available; b) includes a weighting strategy to guide the adversarial loss to focus on samples where the re-ID prediction is highly confident.
     \item We empirically validate the use of local features for cross-modal matching with the proposed architecture inspired by~\cite{sun2018beyond}.
    \item Extensive experiments on three datasets, e.g, SYSU-MM01~\cite{wu2017rgb}, RegDB~\cite{sensors17}, Sketch Re-ID~\cite{pang2018cross} demonstrate the superior performance of our approach. 
\end{itemize}

\section{Related works}
\textbf{Part-level features in person re-ID}.  Employing part-level features for person re-ID was successfully exploited in~\cite{yi2014deep}. They partition the input image into three horizontal strips corresponding to the head, torso, and legs, respectively. The backbone is then applied to each strip individually to extract the part-features.~\cite{sun2018beyond} uniformly partition the final feature map from ResNet50 into six fixed-size horizontal strips and apply independently, a cross-entropy loss to each feature representation.~\cite{zheng2019pyramidal} proposed a multi-stream pyramidal architecture, where each stream generates local features at a different scale. In contrast to the rigid partitioning of the image/feature map, an alternative approach is to generate part-level features with a spatial-attention mechanism. A pioneering work~\cite{zhao2017deeply} incorporates several attention sub-networks to learn a gating map for weighting the final feature representation.
\noindent
\newline \textbf{Infrared-colour person re-ID}. One of the first methods focusing on infrared-colour person re-ID was proposed in~\cite{wu2017rgb}. They introduce a zero-padding layer applied to the input image which promotes the learning of domain-specific filters in their CNN model.~\cite{ijcai18vtreid} incorporates a dual-stream architecture trained with infrared and colour images. As a training strategy, they propose a multi-task loss including a cross-modal triplet and siamese loss term, where the former is being tailored to remove the domain discrepancy and the latter to focus on the re-ID task. Recently,~\cite{DBLP:journals/corr/abs-1907-09659} introduces the skip layer in the model proposed by~\cite{ijcai18vtreid} to fuse the feature representations from the intermediate layers with those from the final layer.~\cite{dai2018cross} attempts to remove the domain discrepancy by incorporating the domain adversarial learning framework~\cite{ganin2016domain}. In addition~\cite{wang2019learning} employs the cycleGAN model~\cite{liu2017unsupervised} to instead remove the domain discrepancy at the pixel-level. Finally,~\cite{basaran2019efficient} proposed an ensemble model including four ResNet152 networks, each individually trained on a different configuration of the input images.

\noindent Similar to~\cite{dai2018cross} we address the domain discrepancy at the feature level. However, inspired by~\cite{wang2019learning} we believe that considering only features from the last layer leads to a sub-optimal solution, therefore we additionally apply the domain adversarial loss to features coming from intermediate layers as they contain significantly more information that is common across the modalities, as shown in our experimental results.
In contrast to all the proposed methods based on the ResNet architecture, we incorporate a variant of the PCB model to extract part-level features, which we prove to be effective for cross-modal matching.

\section{Proposed model}
We introduce a novel framework for infrared-colour person re-ID with two components, i.e, the feature extractor block and the adversarial block.  
Our framework is represented in figure~\ref{diagram_model}. It first extracts features at multiple abstraction levels from intermediate layers of ResNet50 as well as part-level features from the last feature map. With our proposed adversarial learning framework, we aim to remove the domain discrepancy between colour and infrared modalities in the feature space.

\subsection{Feature extractor block} \label{Feature extractor block}
  Our feature extraction block is ResNet50~\cite{DBLP:journals/corr/HeZRS15} with last global average pooling layer (GAP) removed and the extracted feature map partitioned into $n$ horizontal stripes, which is inspired by the PCB approach~\cite{sun2018beyond}.
 Each stripe is passed to an independent embedding block containing a GAP layer followed by a fully connected layer for extracting the part-features. During inference, the final feature representation is obtained by concatenating the part-features from each of the embedding blocks and is defined as follows: $f=[f_1, ..., f_n]$.
\newline
\noindent In contrast to PCB~\cite{sun2018beyond}, our backbone shares the weights between the infrared and colour images only at the final residual module $Resblock_4$. This design choice leads to improved performance in our evaluation. Besides, we extract features with multiple abstraction levels from intermediate layers by applying global average pooling. We use the following notation ${g_j:j=1,..,4}$ to indicate features extracted from ResNet block $j={1,..,4}$. Note that such features are utilized only for adversarial training and are discarded during testing.
%, the motivation for their use is discussed in subsection~\ref{The Domain Adversarial Training Strategy}.

\subsection{Adversarial block}~\label{The Domain Adversarial Training Strategy}
The adversarial block contains a weight evaluation block and a domain classifier. The first computes the entropy of the predicted class distribution corresponding to the current sample. The second utilizes the entropy value to weight the adversarial loss derived from the modality prediction.

\noindent The general training procedure can be summarised as a min-max game, where we alternate between optimizing the domain classifier on a cross-entropy loss term (also known as the domain adversarial loss), and optimising our feature-extraction model as to maximise the same loss. By doing so, the model is learning to remove the domain discrepancy in the extracted feature representations. Besides, we train our feature extraction model to minimise both a cross-entropy and triplet re-ID loss, as to improve the discriminatory capabilities of the learnt feature representation. The multi-task objective function used in our adversarial training framework is:

\vspace{-\baselineskip}
\begin{align} \label{eq2}
    \lambda_{final} = \sum^{n}_{i=1} \lambda_{xent}(f_i) + \lambda_{tri} - \sum_{j=1}^{4} \lambda_{adv}(g_j,w)
\end{align}
\vspace{-\baselineskip}

\noindent where $\lambda_{xent}(f_i)$ denotes the cross-entropy loss term applied to the \ith part-feature $f_i$, $\lambda_{tri}$ corresponds to the batch hard triplet loss \cite{hermans2017defense}, and $\lambda_{adv}(g_j,w)$ is our proposed weighted adversarial loss applied to the intermediate feature representations ${g}_j$ with the weight value $w$.

\noindent Prior to discussing our proposed version of the domain adversarial loss, we briefly introduce the original, vanilla loss function. 

\noindent \textbf{Vanilla domain adversarial loss.} 
\cite{dai2018cross, pang2018cross} defined the original version of domain adversarial loss, which is:

\vspace{-\baselineskip}
\begin{align} \label{eq3}
    \lambda_{adv}(f)= - m \text{log}[\mathcal{D}({{f}})] - (1-m) \text{log}[1-\mathcal{D}({{f}})]
\end{align}
\vspace{-\baselineskip}

\noindent Where $f$ is the feature representation, $m \in \{0,1\}$ is the modality label, where $0$ corresponds to the colour modality and $1$ to the infrared modality; $\mathcal{D}$ is the functional mapping for the domain-classifier which maps the input representation into a single value indicating the probability the input representation comes from  the infrared modality.

\begin{figure}[!b]
\vspace{-\baselineskip}
\centering
\includegraphics[width=1\linewidth, height=1.5in]{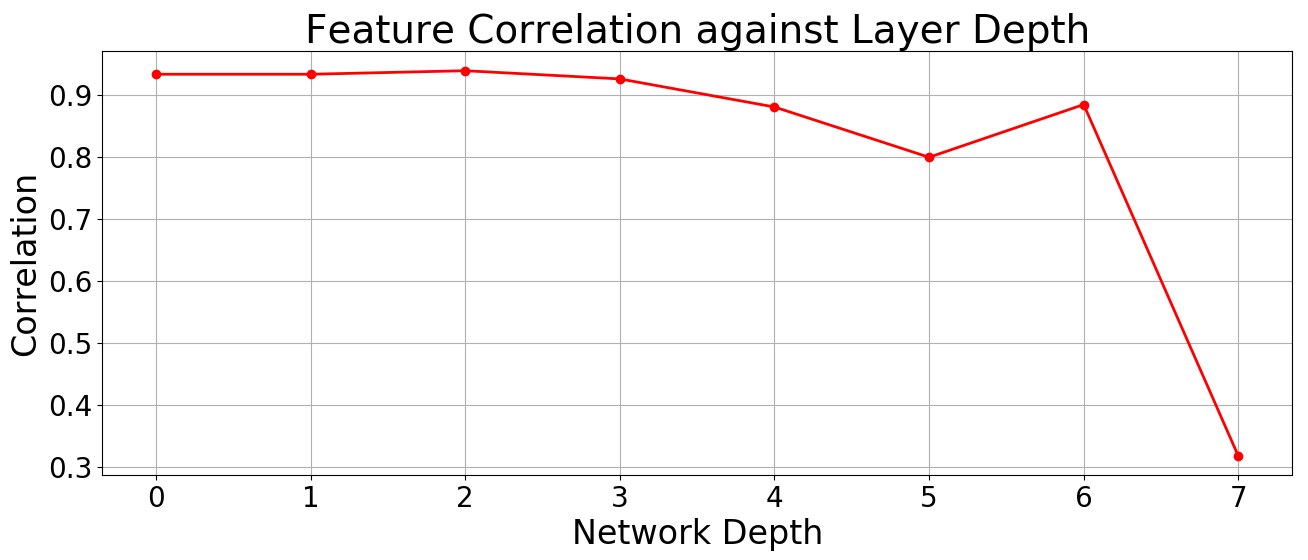}
\vspace{-\baselineskip}
\caption{Average correlation between a set of pairs of features extracted from images containing the same person across two modalities.} 
\label{fig69}
\end{figure}

\noindent \textbf{Proposed weighted adversarial loss.} We propose two novel modifications to the vanilla adversarial loss. 

\noindent The first modification is to apply the adversarial loss to the intermediate representations in our model from the output of the four residual modules in the ResNet50 backbone, as depicted in figure \ref{diagram_model}. This is in contrast to the vanilla configuration which applies the loss term only to the final feature representation. The reasoning behind such a modification comes from our hypothesis that deeply-learned features are susceptible to being modality-specific. In other words, they are distinctive only of one specific modality, i.e, infrared or color. In contrast, the low-level features extracted from the intermediate layers are likely to possess more mutual information across the two modalities.  Thus, applying the domain adversarial loss to shallower layers in our model alleviates the domain discrepancy in the final feature representation output. %meaning less training time and/or samples are required. 
To validate our hypothesis figure~\ref{fig69} illustrates how the correlation between pairs of feature-vectors extracted from the same person across infrared and colour modalities decreases with layer depth in our model.

\noindent The second modification is to weight the adversarial loss term for each input image in the training batch. Inspired by~\cite{long2018conditional} we weight the adversarial loss term for each sample inversely proportional to the entropy in the generated class distribution. The distribution is calculated by averaging the class logit-vectors from each part/stripe output from our feature extraction block, each of which is obtained by taking the part-feature $f_i$ and passing it through the \ith classification layer of our feature extraction block, which is parameterised by $W_i$ (see figure~\ref{diagram_model}). The resultant averaged logit-vector is then softmax normalised to represent a probability distribution.

\noindent The motivation for incorporating this strategy is that the entropy in the generated class distribution for a given input image is inversely proportional to the confidence in id-prediction. For example, if the model produces a high entropy distribution, the confidence in the identity prediction will be low, indicating that the extracted features are not a good representation. To avoid hindering the domain classifier, our strategy gives low importance to such features.   

\noindent The weight generated for a given input sample/image with feature representation $f$ is computed as:

\vspace{-\baselineskip}
\begin{align} \label{eq6}
    w = \frac{1+e^{-H(f)}}{\sum_{k=1}^{M} 1+e^{-H(f^k)}}
\end{align}

\noindent Where $H(f)$ denotes the entropy in the class distribution for the input image with feature representation $f$, and $M$ is the batch size. Specifically, the entropy is computed by:

\vspace{-\baselineskip} 
\begin{align} \label{eq600}
H(f) = -\sum_{j=1}^{C}P(j|f)\text{log}[P(j|f)]
\end{align}

\noindent Where $C$ is the total number of identities and $P(j|f)$ represents the probability of identity $j$ given descriptor $f$.

\begin{figure*}[t!]
\includegraphics[width=0.95\textwidth, height=1.7in]{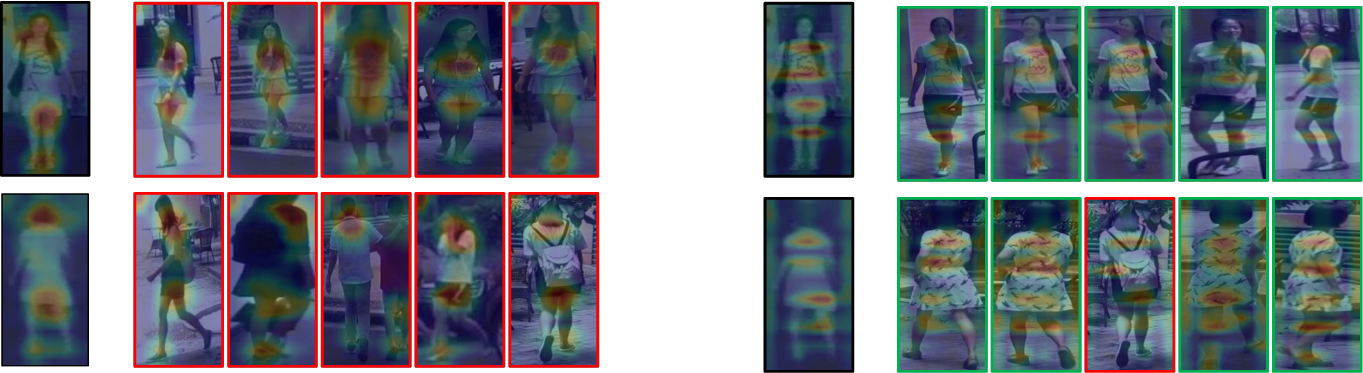}
\caption{ (Left) Example rank-lists for which the ResNet50 model fails. (Right) Corresponding improved rank-lists with our method extracting part-based features from $3$ partitions} 
\label{fig444}
\vspace{-\baselineskip}
\end{figure*}

\newpage
\noindent The weighted domain adversarial loss is thus defined as:

\vspace{-\baselineskip}
\begin{align} \label{eq10101}
    \lambda_{adv}(g,w)= \sum_{j=1}^4 \lambda_{adv}(g_{j},w)
\end{align}
\vspace{-\baselineskip}

\vspace{-\baselineskip}
\begin{align} \label{eq10101}
    \lambda_{adv}(g_{j},w)=-w(m\cdot\text{log}[\mathcal{D}(g_{j})]+ \nonumber (1-m)\cdot\text{log}[1-\mathcal{D}(g_j)])
\end{align}
\vspace{-\baselineskip}

\noindent where ${g}_{j}$ denotes the feature-vector extracted from \jth residual module.
   
\section{Experimental settings}
\subsection{Datasets}
 
%The Sketch Re-ID and RegDB datasets are experimented with only in section \ref{Comparison With State-Of-The-Art}.

\noindent\textbf{SYSU-MM01} contains 491 identities, with images collected across six different cameras on a university campus in both an outdoor and indoor environment. Two of the cameras capturing infrared and the rest capturing colour images. We follow the all-search, single-shot evaluation protocol defined in~\cite{wu2017rgb}.
    
\noindent\textbf{RegDB} contains 412 identities, with each person captured in an outdoor environment from the same camera view in both thermal and colour modalities. For evaluation, we follow the protocol defined in~\cite{sensors17}.
    
\noindent\textbf{Sketch Re-ID}  contains only 200 identities with each person viewed across two different cameras in an outdoor environment. Furthermore, a sketch-artist produces a single sketch for each person based on a description given by volunteer witnesses. We follow the evaluation protocol defined in~\cite{pang2018cross}. 
\subsection{Implementation details} \label{Training set-up}
\textbf{Our approach.} In the feature extractor block the dimension for each extracted part-feature is $512$, and we use only $n=3$ partitions/stripes. The ResNet50 backbone has been pre-trained on ImageNet~\cite{krizhevsky2012imagenet}. During training as a pre-processing step, all images are resized to dimension $288\times144$ and are normalized. Images are augmented at random, by horizontally flipping and the removal of blocks (random erasing)~\cite{zhong2017random}, with the respective probabilities $0.5$ and $0.2$. Furthermore, to avoid over-fitting we combine the training set with Market-1501~\cite{zheng2015scalable}. We randomly split it in half based on camera-view and for one half we only use the red channel of the RGB images, such that they appear perceptually similar to infrared. We train our architecture for 35 epochs with our proposed multi-task loss function. However, the total training time is 70 epochs as we alternate between optimizing the feature extractor block and the domain classifier. The optimization algorithm used is Stochastic Gradient Descent with momentum, were the learning-rate for the PCB model and domain classifier is $0.01$, however for the pre-trained weights in the ResNet50 backbone the learning rate is $0.001$.

\noindent \textbf{Baseline.} The baseline for experiments is our feature extractor block, trained only on the cross-entropy and triplet loss terms, that is without the contributions discussed in section~\ref{The Domain Adversarial Training Strategy}. Otherwise, the training set-up is identical to that specified for our approach above.

\begin{table*}[!t]
\vspace{-2\baselineskip}
\renewcommand{\arraystretch}{0.9}
\captionof{table}{Results on the SYSU-MM01 and RegDB datasets.\vspace{-0.5\baselineskip}}
\centering
\begin{tabular}{ p{5.5cm}|p{1.5cm}|p{1.5cm}|p{1.5cm}||p{1.5cm}|p{1.5cm}|p{1.5cm}}
\toprule
\multirow{2}{*}{Methods} &  \multicolumn{3}{c||}{RegDB} & \multicolumn{3}{c}{SYSU-MM01}\\ \cline{2-7}
{} & rank-1 & rank-10 & mAP & rank-1 & rank-10 & mAP \\
\hline
Zero-Padding \cite{wu2017rgb}  & 17.8 & 34.2 & 18.9 & 14.8 & 54.1 & 16.0 \\ \hline
BCTR \cite{ijcai18vtreid} & 32.7 & 57.6 & 31.0 & 16.1 & 54.9  & 19.2 \\ \hline
cmGAN \cite{dai2018cross} & - & - & - & 27.0 & 67.5 & 27.8 \\ \hline
$D^{2}$RL \cite{wang2019learning} & 43.4 & 66.1 & 44.1 & 28.9 & 70.6 & 29.2 \\ \hline
EDFL \cite{DBLP:journals/corr/abs-1907-09659} & 52.6 & 72.1 & 53.0 & 37.0 & 84.5 & 40.8 \\ \hline
LZM \cite{basaran2019efficient} & - & - & - & 48.9 & 90.7 & 50.0 \\ \hhline{=======}
Baseline & 58.8 & 79.7  & 56.5 & 51.5 & 89.3 & 50.4 \\ \hline
\textbf{adversarial loss shallow + weighting} & \textbf{61.6} & \textbf{82.7} & \textbf{59.0} & \textbf{55.2} & \textbf{90.9}  & \textbf{53.2} \\
\bottomrule
\end{tabular}
\vspace{-0.5\baselineskip}
\label{table1}
\vspace{-\baselineskip}
\end{table*}

\section{Experiments}
We first present results that demonstrate the benefit of using part-based feature and the proposed adversarial loss.
These results are reported in sections~\ref{Effectiveness Of PCB For Cross-Modal Matching} and~\ref{Ablation Study On The Domain Adversarial Loss} have been obtained by evaluating on the SYSU-MM01.
Finally, we compare our method to state-of-the-art works.

\begin{figure}[!b] 
\centering
\includegraphics[width=1.0\linewidth, height=1.75in]{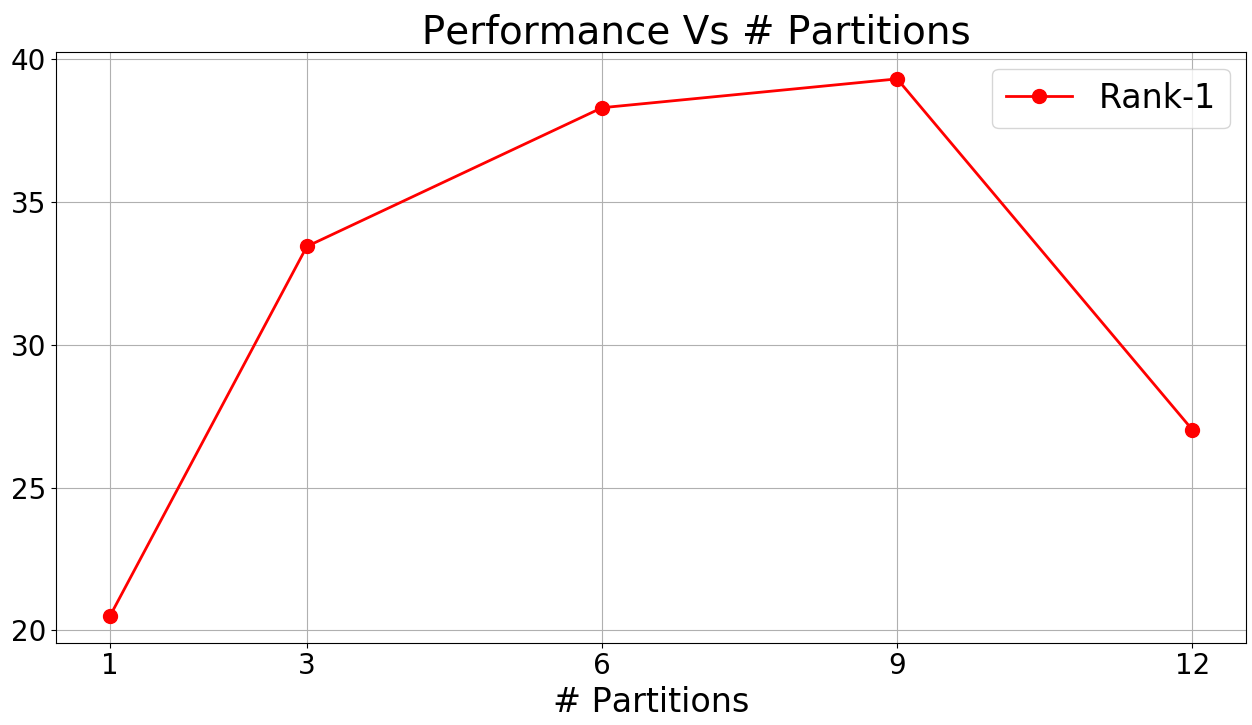}
\vspace{-1\baselineskip}
\caption{Rank-1 accuracy against number of partitions in feature extractor block.} 
\label{fig5}
\end{figure}

\subsection{Effectiveness of extracting part-based features} \label{Effectiveness Of PCB For Cross-Modal Matching}

\begin{table}[!t]
\smallskip
\smallskip
\centering
\captionof{table}{Ablation study on proposed weighted adversarial loss}
\vspace{-0.5\baselineskip}
\renewcommand{\arraystretch}{0.9}
\begin{tabular}{ c|c|c }
\toprule
Methods &  rank-1 & mAP \\ \hline
Baseline & 51.5 & 50.4 \\ \hline
vanilla adversarial loss & 48.8 & 48.8 \\ \hline
adversarial loss shallow & 51.9 &  51.3 \\ \hline
\textbf{adversarial loss shallow+weighting} & \textbf{55.2} & \textbf{53.2}\\
\bottomrule
\end{tabular}
\label{table2}
\vspace{-1.5\baselineskip}
\end{table}

The effect of using a number of partitions/part-features in the feature extractor block of our baseline is demonstrated in figure~\ref{fig5}.  Using the model with nine partitions gives optimal performance, providing a boost close to $20\%$ in rank-1 accuracy compared to using one partition only. For the latter configuration, the feature extractor block in our baseline is identical to the original ResNet50 network with a GAP layer~\cite{DBLP:journals/corr/HeZRS15}. Which is a commonly used architecture in cross-modal person re-ID, where the model extracts only a single, global-level feature-vector representation from the input image. Note that in this experiment random erasing had not been applied, nor had we utilized the training images from the Market-1501 dataset to make the comparison fair. When such techniques are incorporated during training we observe the feature extractor block with three partitions results in the best performance for our baseline. 
\noindent The performance boost when using multiple part-features, we attribute to our training regime as presented in section~\ref{The Domain Adversarial Training Strategy}, whereby a cross-entropy loss is applied to each extracted part-feature.
This is because the receptive field associated with each extracted part-feature corresponds to small, non-overlapping, horizontal stripes of the input image, each of which contains limited discriminative information. Hence, by optimizing each part on its cross-entropy loss, the backbone network is forced to focus on all salient regions in each corresponding image-stripe.
Thus we can attend to regions on a person not originally seen when using the ResNet50 architecture in our baseline. To better illustrate this, figure~\ref{fig444} provides example rank-lists where the baseline ResNet50 fails (Left), and how such failure cases are solved with our feature extractor block with three partitions (Right). Furthermore, the respective Class Activation Maps (CAMs)~\cite{zhou2016learning} have been overlayed on the images, to illustrate the parts of the image which the respective networks are attending too. From the top-left rank-list, we can observe that the ResNet50 model is only able to focus on the person's face and shorts. Such information is insufficient to correctly match the relevant gallery images to the probe image in the presence of other similar looking identities in the gallery set. However, from observing the top-right rank-list we can see that our model with three partitions can additionally attend to a part of the logo on the persons t-shirt. By exploiting such a unique region on the person, our approach is now able to correctly match the probe image to the relevant gallery images. 

\subsection{Ablation study on our domain adversarial loss} \label{Ablation Study On The Domain Adversarial Loss}
Table~\ref{table2} empirically validates the benefits of our two proposed modifications. The method "vanilla adversarial loss" corresponds to the configuration where we only apply the adversarial loss to deep layers in our model, specifically to the feature-vectors extracted from the final layer and the output from the ResNet50 backbone, similar~\cite{dai2018cross, pang2018cross}. The results in table~\ref{table2} show that such a configuration performs worse than the baseline. The configuration "adversarial loss shallow" corresponds to the case when we additionally applied the adversarial loss to shallow layers in our model. Specifically, we apply the loss term to the output feature-maps from Resblock1,2,3 and 4 in the ResNet50 backbone. Such a configuration provides a performance boost over the baseline, which suggests that attempting to remove the domain discrepancy at an early stage in the network is indeed beneficial. Furthermore, when we also incorporate our proposed weighting strategy to each adversarial loss we can further improve the performance of our method.
\subsection{Comparison with state-of-the-art} \label{Comparison With State-Of-The-Art}
Tables \ref{table1} and \ref{table3} report the results of notable previous works on the SYSU-MM01, RegDB and Sketch Re-ID datasets.  Our strong baseline model can consistently outperform the previous state-of-the-art across all datasets. Furthermore, by additionally incorporating our novel domain adversarial loss, we can improve the performance further by a notable margin. The presented results validate our proposed contributions not only for infrared-colour person re-ID but also for the tasks of thermal-colour and sketch-colour person re-ID. This shows that our approach is effective for learning discriminative, non-colour related features.
\begin{table}[!t]
\smallskip
\smallskip
\centering
\captionof{table}{Results on the Sketch Re-ID dataset.}
\vspace{-0.5\baselineskip}
\renewcommand{\arraystretch}{0.9}
\begin{tabular}{ c|c|c|c }
\toprule
Methods & rank-1 & rank-10\\
\hline
HOG+LBP+rankSVM  & 5.1 & 28.30\\
\hline
GN Siamese \cite{sangkloy2016sketchy} & 28.9 &  62.4  \\
\hline
AFL \cite{pang2018cross} & 34.0 & 72.50 \\
\hhline{====}
Baseline  & 56.7 & 87.3 \\
\hline
\textbf{adversarial loss shallow+weighting} & \textbf{62.2} & \textbf{91.4} \\
\bottomrule
\end{tabular}
\label{table3}
\vspace{-1.5\baselineskip}
\end{table}

\section{Conclusions}
We have proposed a novel approach to cross-modal person re-ID which achieved state-of-the-art performance across three popular cross-modal re-id datasets. Furthermore, through extensive experimentation we empirically justified each of our contributions. We were able to demonstrate that incorporating a part based model brings significant improvements, in comparison to using the conventional ResNet50 network. In addition, we validated our two novel modifications to the generic domain adversarial loss~\cite{ganin2016domain}.

\section*{Acknowledgments}
This research was supported by EPSRC FACER2VM  EP/N007743/1.

\bibliography{final_paper_V3}

\begin{thebibliography}{10}
\providecommand{\url}[1]{#1}
\csname url@samestyle\endcsname
\providecommand{\newblock}{\relax}
\providecommand{\bibinfo}[2]{#2}
\providecommand{\BIBentrySTDinterwordspacing}{\spaceskip=0pt\relax}
\providecommand{\BIBentryALTinterwordstretchfactor}{4}
\providecommand{\BIBentryALTinterwordspacing}{\spaceskip=\fontdimen2\font plus
\BIBentryALTinterwordstretchfactor\fontdimen3\font minus
  \fontdimen4\font\relax}
\providecommand{\BIBforeignlanguage}[2]{{%
\expandafter\ifx\csname l@#1\endcsname\relax
\typeout{** WARNING: IEEEtran.bst: No hyphenation pattern has been}%
\typeout{** loaded for the language `#1'. Using the pattern for}%
\typeout{** the default language instead.}%
\else
\language=\csname l@#1\endcsname
\fi
#2}}
\providecommand{\BIBdecl}{\relax}
\BIBdecl

\bibitem{ganin2016domain}
Y.~Ganin, E.~Ustinova, H.~Ajakan, P.~Germain, H.~Larochelle, F.~Laviolette,
  M.~Marchand, and V.~Lempitsky, ``Domain-adversarial training of neural
  networks,'' \emph{The Journal of Machine Learning Research}, 2016.

\bibitem{wu2017rgb}
A.~Wu, W.-S. Zheng, H.-X. Yu, S.~Gong, and J.~Lai, ``Rgb-infrared
  cross-modality person re-identification,'' in \emph{Proceedings of the IEEE
  international conference on computer vision}, 2017.

\bibitem{dai2018cross}
P.~Dai, R.~Ji, H.~Wang, Q.~Wu, and Y.~Huang, ``Cross-modality person
  re-identification with generative adversarial training.'' in \emph{IJCAI},
  2018.

\bibitem{wang2019learning}
Z.~Wang, Z.~Wang, Y.~Zheng, Y.-Y. Chuang, and S.~Satoh, ``Learning to reduce
  dual-level discrepancy for infrared-visible person re-identification,'' in
  \emph{CVPR}, 2019.

\bibitem{sun2018beyond}
Y.~Sun, L.~Zheng, Y.~Yang, Q.~Tian, and S.~Wang, ``Beyond part models: Person
  retrieval with refined part pooling (and a strong convolutional baseline),''
  in \emph{ECCV}, 2018.

\bibitem{sensors17}
D.~T. Nguyen, H.~G. Hong, K.~W. Kim, and K.~R. Park, ``Person recognition
  system based on a combination of body images from visible light and thermal
  cameras,'' \emph{Sensors}, 2017.

\bibitem{pang2018cross}
L.~Pang, Y.~Wang, Y.-Z. Song, T.~Huang, and Y.~Tian, ``Cross-domain adversarial
  feature learning for sketch re-identification,'' in \emph{ACM}, 2018.

\bibitem{yi2014deep}
D.~Yi, Z.~Lei, S.~Liao, and S.~Z. Li, ``Deep metric learning for person
  re-identification,'' in \emph{ICPR}, 2014.

\bibitem{zheng2019pyramidal}
F.~Zheng, C.~Deng, X.~Sun, X.~Jiang, X.~Guo, Z.~Yu, F.~Huang, and R.~Ji,
  ``Pyramidal person re-identification via multi-loss dynamic training,'' in
  \emph{CVPR}, 2019.

\bibitem{zhao2017deeply}
L.~Zhao, X.~Li, Y.~Zhuang, and J.~Wang, ``Deeply-learned part-aligned
  representations for person re-identification,'' in \emph{ICCV}, 2017.

\bibitem{ijcai18vtreid}
M.~Ye, Z.~Wang, X.~Lan, and P.~C. Yuen, ``Visible thermal person
  re-identification via dual-constrained top-ranking,'' in \emph{IJCAI}, 2018.

\bibitem{DBLP:journals/corr/abs-1907-09659}
H.~Liu and J.~Cheng, ``Enhancing the discriminative feature learning for
  visible-thermal cross-modality person re-identification,'' \emph{CoRR}, 2019.

\bibitem{liu2017unsupervised}
M.-Y. Liu, T.~Breuel, and J.~Kautz, ``Unsupervised image-to-image translation
  networks,'' in \emph{NeurIPS}, 2017.

\bibitem{basaran2019efficient}
E.~Basaran, M.~Gokmen, and M.~E. Kamasak, ``An efficient framework for
  visible-infrared cross modality person re-identification,'' \emph{arXiv
  preprint arXiv:1907.06498}, 2019.

\bibitem{DBLP:journals/corr/HeZRS15}
K.~He, X.~Zhang, S.~Ren, and J.~Sun, ``Deep residual learning for image
  recognition,'' in \emph{CVPR}, 2016.

\bibitem{hermans2017defense}
A.~Hermans, L.~Beyer, and B.~Leibe, ``In defense of the triplet loss for person
  re-identification,'' \emph{arXiv preprint arXiv:1703.07737}, 2017.

\bibitem{long2018conditional}
M.~Long, Z.~Cao, J.~Wang, and M.~I. Jordan, ``Conditional adversarial domain
  adaptation,'' in \emph{NeurIPS}, 2018.

\bibitem{krizhevsky2012imagenet}
A.~Krizhevsky, I.~Sutskever, and G.~E. Hinton, ``Imagenet classification with
  deep convolutional neural networks,'' in \emph{NeurIPS}, 2012.

\bibitem{zhong2017random}
Z.~Zhong, L.~Zheng, G.~Kang, S.~Li, and Y.~Yang, ``Random erasing data
  augmentation,'' \emph{arXiv preprint arXiv:1708.04896}, 2017.

\bibitem{zheng2015scalable}
L.~Zheng, L.~Shen, L.~Tian, S.~Wang, J.~Wang, and Q.~Tian, ``Scalable person
  re-identification: A benchmark,'' in \emph{Proceedings of the IEEE
  international conference on computer vision}, 2015.

\bibitem{zhou2016learning}
B.~Zhou, A.~Khosla, A.~Lapedriza, A.~Oliva, and A.~Torralba, ``Learning deep
  features for discriminative localization,'' in \emph{CVPR}, 2016.

\bibitem{sangkloy2016sketchy}
P.~Sangkloy, N.~Burnell, C.~Ham, and J.~Hays, ``The sketchy database: learning
  to retrieve badly drawn bunnies,'' \emph{ACM Transactions on Graphics}, 2016.

\end{thebibliography}

\end{document}